\documentclass[11pt, a4paper, logo]{lumia}

\usepackage[numbers,sort&compress]{natbib}
\definecolor{infrecred}{HTML}{D2071E}
\definecolor{abstrabg}{HTML}{FCDDDC}
\colorlet{abstradark}{infrecred}
\definecolor{reportaccent}{HTML}{BB0819}
\colorlet{reporttoc}{reportaccent}
\definecolor{cayenne}{HTML}{941100}
\hypersetup{allcolors=cayenne}
\definecolor{darkblue}{rgb}{0, 0, 0.5}
\definecolor{lightblue}{HTML}{D0CFEA}

\usepackage{algorithm}
\usepackage{algpseudocode}
\usepackage{multirow}
\usepackage{xurl}

\newcommand{\subfigref}[2]{\hyperref[#1]{\ref*{#1}(#2)}}

\graphicspath{{fig/}}

\setheadertext{AB-Attention}

\title{%
  AB-Attention: Accurate Block Attention for Accelerating Video Diffusion Transformers
}

\setheadertitle{AB-Attention: Accurate Block Attention for Accelerating Video Diffusion Transformers}

\renewcommand\Affilfont{\normalfont\fontsize{9pt}{11pt}\selectfont}
\author{
  \shortstack{
    Bin Xu\textsuperscript{1,2}\quad Mo Zhang\textsuperscript{3}\quad Zhengfeiyang Zhang\textsuperscript{1,4}\quad Xinfeng Yang\textsuperscript{5}\quad Yixing Ma\textsuperscript{1,6}\authorcr
    Wei Wang\textsuperscript{7}\quad HeXin Wang\textsuperscript{1}\quad Pengfei Zhou\textsuperscript{1,8\textdagger}\quad Wangbo Zhao\textsuperscript{7\textdagger}\authorcr
    {\Affilfont\rule{0pt}{12pt}\textsuperscript{1}InfRec\quad \textsuperscript{2}The Chinese University of Hong Kong, Shenzhen\quad \textsuperscript{3}Shandong University}\authorcr
    {\Affilfont\textsuperscript{4}The University of Tokyo\quad \textsuperscript{5}Fudan University\quad \textsuperscript{6}University of California, Berkeley}\authorcr
    {\Affilfont\textsuperscript{7}The Hong Kong University of Science and Technology\quad \textsuperscript{8}National University of Singapore\qquad \textsuperscript{\textdagger}Corresponding authors}
  }
}
\hypersetup{
  pdftitle={AB-Attention: Accurate Block Attention for Accelerating Video Diffusion Transformers},
  pdfauthor={Bin Xu, Mo Zhang, Zhengfeiyang Zhang, Xinfeng Yang, Yixing Ma, Wei Wang, HeXin Wang, Pengfei Zhou, Wangbo Zhao}
}

\begin{document}

\begin{abstract}
Video diffusion transformers generate high-quality videos, but the quadratic
cost of 3D full attention dominates their inference. Dynamic block-sparse
attention reduces this cost by computing exact attention only on key/value
blocks selected from current activations. However,
its fidelity degrades under aggressive sparsity: blocks are scored from coarse
summaries or formed by costly global clustering, pruned blocks vanish from the
softmax, and whole tiles are kept or dropped even when only a few of their keys
matter. To address these limitations, we propose Accurate Block Attention (\textbf{AB-Attention}), a
training-free method that accurately approximates dense attention at high
sparsity. Its Intra-Block Clustering
Selector (IBCS) estimates block
importance faithfully yet cheaply by clustering queries only within each
query block, without permuting tokens. Sampling with Inclusion-Probability
Weighting (SIP) then samples
blocks in proportion to these estimates and reweights them by inverse
inclusion probability, so that one pass estimates dense attention rather than
truncating it. Finally, its kernel gathers key/value blocks much smaller than a
tile into full tensor-core tiles, refining selection at no extra tensor-core
cost. On Wan2.1 and HunyuanVideo, AB-Attention skips 89--92\% of attention,
whereas the baselines compute 1.8--3.6$\times$ as much, yet it still improves PSNR
by up to 4.4~dB over the strongest baseline and achieves up to 2.23$\times$
end-to-end speedup. Our code is available at
\url{https://github.com/InfRec-Cardinal/AB-Attention}.

\end{abstract}

\teaserfigure{%
\vskip-8pt
\begin{minipage}{\linewidth}
\centering
\includegraphics[width=0.88\linewidth]{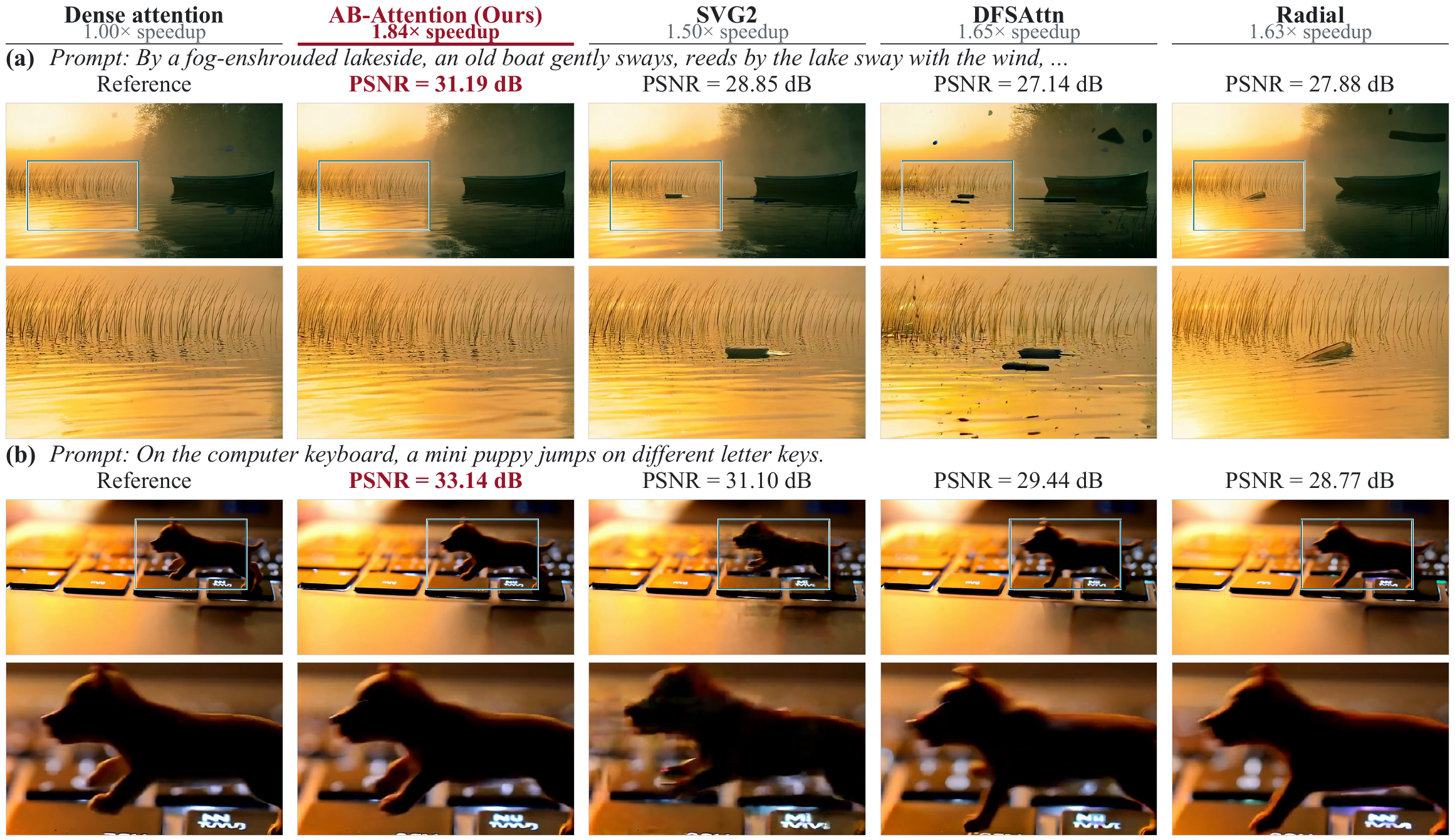}
{\captionsetup{font=small,hypcap=false}
\captionof{figure}{Examples on Wan2.1-T2V-14B, with one frame and an enlarged region per
  method. AB-Attention is closest to dense attention in full-video PSNR while
  achieving the highest speedup.}
\label{fig:teaser}}
\end{minipage}
\vskip4pt}

\maketitle

\clearpage
\begingroup
\hypersetup{linkcolor=reporttoc,linktoc=section}
\setlength{\parskip}{0pt}
\setcounter{tocdepth}{3}
\tableofcontents
\endgroup
\clearpage

\section{Introduction}
\label{sec:introduction}

Video diffusion transformers~\citep{wan2025wan,kong2024hunyuanvideo} primarily rely on 3D full attention to capture
spatial and temporal dependencies, enabling high-quality
generation. However, the computational
cost of full attention grows quadratically with the sequence length.
As the video resolution and number of frames increase,
attention becomes increasingly expensive. Since this
computation is repeated across transformer layers and denoising steps,
attention has long been a major inference bottleneck
\citep{xi2025sparsevideogen,zhang2025sta}. Reducing the cost of
attention while preserving generation quality is therefore essential
for generating longer and higher-resolution videos.

Sparse attention reduces this cost by computing only a subset of
query--key pairs. Early methods apply predefined spatial or temporal
patterns, such as local 3D windows~\citep{zhang2025sta}, energy-decay
masks~\citep{li2025radial}, or per-head spatial and temporal
masks~\citep{xi2025sparsevideogen}. These predefined layouts reduce the need
for online block scoring, but their limited adaptability to attention variations
across prompts, heads, and denoising steps can require retaining many
unimportant pairs to preserve important interactions.
Dynamic methods instead select
key/value blocks from the current activations, adapting the sparsity pattern
to the input at the cost of online scoring. Since block-sparse
kernels execute attention in fixed-size tiles, they partition queries and keys into tile-aligned
blocks, estimate block importance cheaply from mean-pooled
tokens~\citep{zhang2025spargeattention} or antidiagonal attention
samples~\citep{xu2025xattention}, and compute exact attention only on
the most important blocks. SVG2~\citep{yang2025svg2} instead forms the blocks
by clustering and permuting tokens.


Despite this progress, dynamic methods remain limited by three design
choices: \textbf{1) Importance estimates are coarse or costly.} Pooling a
block into one representative mixes tokens with different attention patterns
and can misrank blocks, whereas clustering yields faithful representatives
only through sequence-wide $k$-means and token
permutation~\citep{yang2025svg2}. \textbf{2) Top-$k$ selection truncates
attention.} Discarding a block removes its contribution from both the attention
numerator and the softmax normalizer, which better scores cannot recover.
Correction-based
methods~\citep{liu2025rectifiedspaattn,li2026pisa,zhou2026svgear} approximate
the discarded remainder, but add a separate approximation branch to the exact
sparse computation. \textbf{3) The selection unit is the execution tile.} A
tile-sized key block is kept or discarded as a whole even when only a few of
its keys matter, and reordering tokens makes blocks more coherent without
changing this unit. Finer units need not slow execution, as a kernel can
gather them into full tiles, but they multiply the candidates to score.

We therefore propose Accurate Block Attention (\textbf{AB-Attention}), a
training-free method that directly addresses these limitations. The \emph{Intra-Block
Clustering Selector} (IBCS) scores key/value blocks with lightweight query clusters formed
only within each query block, avoiding costly global clustering and token permutation.
\emph{Sampling with Inclusion-Probability Weighting} (SIP) samples blocks in
proportion to these scores and reweights them by inverse inclusion
probability. Every block can thus contribute, and a single sparse attention pass
yields unbiased estimates of both the attention numerator and the softmax normalizer.
AB-Attention further performs \emph{fine-grained selection with tiled
execution}: it selects key/value blocks much smaller than an execution tile
and gathers them into full tensor-core tiles in its kernel, without reordering any tokens.
The sampling plans are reused across a few nearby denoising steps, while attention is
recomputed at every step~\citep{xia2025adaspa}.



Extensive experiments on representative video diffusion transformers, including Wan2.1~\citep{wan2025wan} and HunyuanVideo~\citep{kong2024hunyuanvideo}, demonstrate that \mbox{AB-Attention} significantly advances the quality--efficiency frontier of training-free sparse attention. It skips
\mbox{89--92\%} of attention computation, whereas the competing baselines compute
1.8--3.6$\times$ as much, and delivers
1.84--2.23$\times$ end-to-end speedups on a single H20 GPU, while achieving the highest fidelity to
dense attention and improving PSNR by 1.6--4.4~dB over the strongest
baselines, including SVG2~\citep{yang2025svg2} and DFSAttn~\citep{hu2026dfsattn}. The examples in Figure~\ref{fig:teaser} visually illustrate these fidelity gains.

\section{Preliminaries}
\label{sec:preliminaries}

\begin{figure}[t]
  \centering
  \setlength{\abovecaptionskip}{4pt}
  \includegraphics[width=\linewidth]{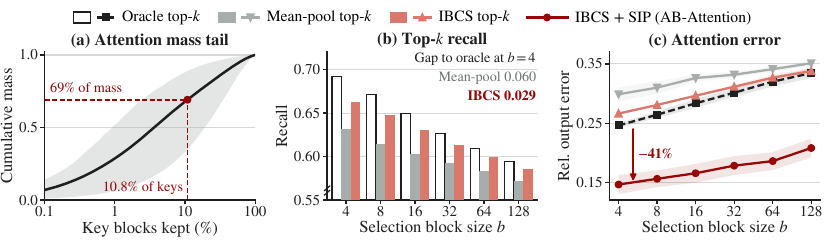}
  \caption{\textbf{Attention mass, recall, and error on Wan2.1-T2V-14B} at
  89.16\% sparsity.
  (a)~Sorted cumulative attention mass of 4-token key blocks; the band
  spans the 10th--90th percentiles of query blocks.
  (b)~Top-$k$ recall: the fraction of each query block's dense attention
  mass covered by the selected key blocks. Finer blocks keep more mass, and IBCS
  halves the gap of mean-pool to the oracle at $b{=}4$.
  (c)~Relative $\ell_2$ error of the attention output against dense
  attention. Sampling with SIP on IBCS scores lowers the error 41\% below oracle
  top-$k$ at $b{=}4$.}
  \label{fig:granularity}
\end{figure}
\paragraph{Dense attention.}
Video diffusion transformers apply self-attention to tokens across space and
time. Given $n$ tokens, let $Q,K,V\in\mathbb R^{n\times d}$ denote the query,
key, and value matrices for one attention head. Dense attention computes
\begin{equation}
  O=\operatorname{softmax}\!\left(\frac{QK^\top}{\sqrt d}\right)V.
  \label{eq:dense_attention}
\end{equation}
Its $\mathcal O(n^2d)$ complexity makes dense attention costly
for high-resolution videos.
FlashAttention~\citep{dao2022flashattention} accelerates exact attention by
reducing data movement and intermediate storage, but still computes all
query--key interactions, leaving the arithmetic cost quadratic in sequence length.

\paragraph{Sparse attention.}
Sparse attention reduces this arithmetic by computing only selected
query--key pairs. Conceptually, it adds a mask
$\mathcal M\in\{-\infty,0\}^{n\times n}$ to the attention logits:
\begin{equation}
  O_{\mathrm{sparse}}=
  \operatorname{softmax}\!\left(\frac{QK^\top}{\sqrt d}+\mathcal M\right)V.
  \label{eq:sparse_attention}
\end{equation}
An entry $\mathcal M_{uv}=0$ keeps the interaction between query $u$ and key
$v$, while $\mathcal M_{uv}=-\infty$ skips it. Block-sparse attention makes
this pattern hardware-friendly by partitioning the queries, keys, and values
into blocks of $B$ tokens. It keeps or skips each query--key block pair as a
whole, so every retained pair can be evaluated using dense matrix operations.
The block size determines a trade-off: larger blocks improve hardware
utilization but select more coarsely, whereas smaller blocks provide finer
selection but may reduce execution efficiency.

\section{Methodology}
\label{sec:method}

\begin{figure}[t]
  \centering
  \includegraphics[width=\linewidth]{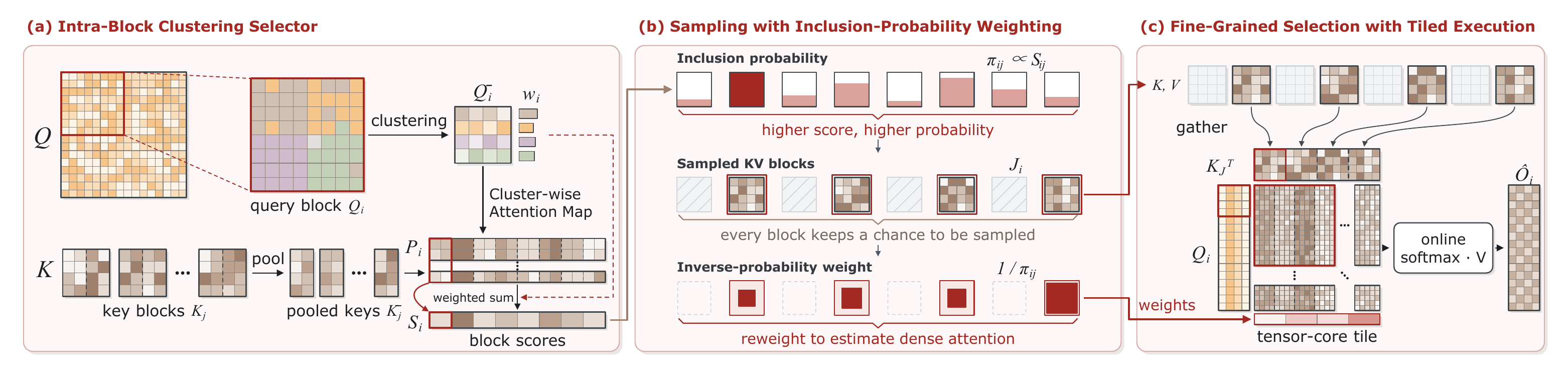}
  \caption{\textbf{Overview of AB-Attention.}
  (a)~IBCS clusters each query block $Q_i$ into centroids
  $\bar Q_i$ with weights $w_i$, and scores every key/value block by the
  weighted attention of the centroids to its pooled keys.
  (b)~SIP samples each block with probability $\pi_{ij}$ proportional to its
  score $S_{ij}$, so no block is excluded, and weights it by $1/\pi_{ij}$ to
  estimate full attention.
  (c)~The kernel gathers the sampled blocks into full tensor-core tiles
  and attends to them with logits shifted by $-\log\pi_{ij}$.}
  \label{fig:ab_overview}
\end{figure}

As illustrated in Figure~\ref{fig:ab_overview}, AB-Attention selects key/value
blocks much smaller than an execution tile and estimates the dense attention of
each query block from a weighted sample of them. Section~\ref{sec:selector}
introduces the Intra-Block Clustering Selector (IBCS), which scores these blocks
with query clusters formed within each query block. Section~\ref{sec:sip}
introduces Sampling with Inclusion-Probability Weighting (SIP), which samples
blocks in proportion to these scores and reweights them by inverse inclusion
probability, so that one block-sparse pass estimates dense attention rather
than truncating it. Section~\ref{sec:descriptor_execution} introduces
fine-grained selection with tiled execution: its kernels run these small blocks
on full tensor-core tiles without reordering tokens, preserving the efficiency
of block-sparse attention. Algorithm~\ref{alg:ab_attention} summarizes the
method for one attention head at one denoising step.

\subsection{Intra-Block Clustering Selector (IBCS)}
\label{sec:selector}

We divide the queries into blocks $Q_i\in\mathbb R^{B_q\times d}$ of $B_q$
tokens, and the keys and values into smaller blocks
$K_j,V_j\in\mathbb R^{b\times d}$ of $b$ tokens. This asymmetry is deliberate:
smaller key/value blocks make selection finer, while query blocks remain aligned
with execution tiles. As Figure~\subfigref{fig:granularity}{b} shows, shrinking
the key/value blocks from 128 to 4 tokens raises the attention mass retained by
oracle top-$k$ from 59\% to 69\%. Section~\ref{sec:descriptor_execution}
explains how the kernel packs sampled blocks into full execution tiles, allowing
finer selection without increasing attention computation at a fixed budget. Smaller
query blocks offer a different trade-off: query blocks sharing an execution tile
must compute the union of their selected key/value blocks, while running them in
smaller tiles can underutilize tensor cores. We therefore keep each query block
aligned with the execution-tile height and shrink only the key/value blocks.

To score the key/value blocks for $Q_i$, IBCS needs a compact summary of its
$B_q$ queries. Mean-pooling the queries across the full block~\citep{zhang2025spargeattention}
or its positional sub-blocks~\citep{hu2026dfsattn} is inexpensive, but can blur
the distinct attention preferences of different queries. Clustering tokens over
the whole sequence~\citep{yang2025svg2} can keep such queries apart, but requires
$k$-means over all tokens and reorders them, which is costly for long,
high-resolution videos. IBCS instead clusters queries within each query block and
uses the clusters only for scoring; attention itself still runs in the original
token order.

\paragraph{Intra-block query clustering.}
IBCS partitions the queries of $Q_i$ into $R$ clusters
$\mathcal C_{i1},\dots,\mathcal C_{iR}$ with a single \mbox{$k$-means}
iteration: starting from $R$ well-separated queries as initial centers, it
assigns each query to its nearest center. Each cluster $\mathcal C_{ir}$ is summarized by
its centroid $\bar q_{ir}$, the mean of its member queries, and its weight
$w_{ir}$, the fraction of the queries of $Q_i$ that it contains. Stacking them over all $R$ clusters gives the
full centroid matrix $\bar Q_i\in\mathbb R^{R\times d}$ and the corresponding weight vector
$w_i\in\mathbb R^{R}$.

\paragraph{Block scores.}
For each query block, IBCS scores all $M=n/b$ key/value blocks of the
sequence. Each block holds only $b$ tokens, so the keys need far less
summarization than the queries: IBCS only mean-pools small groups of adjacent
keys, as detailed in Appendix~\ref{app:representatives}. $\bar K_j$ denotes the pooled keys
of block $K_j$, and $\bar K=[\bar K_1;\dots;\bar K_M]$ stacks those of all
blocks. The
score of block $j$ is the attention mass on its pooled keys, averaged over the
centroids with the cluster weights:
\begin{equation}
  S_{ij}=\sum_{r=1}^{R}w_{ir}\,
  \frac{\sum_{\bar k\in\bar K_j}\exp\!\left(\bar q_{ir}\bar k^\top/\sqrt d\right)}
       {\sum_{\bar k\in\bar K}\exp\!\left(\bar q_{ir}\bar k^\top/\sqrt d\right)},
  \label{eq:block_score}
\end{equation}
where the fraction is the softmax attention that centroid $\bar q_{ir}$ assigns
to the pooled keys of block $j$. A block that only one cluster attends to thus
still keeps a score in proportion to that cluster's weight, whereas a single mean
query of the whole block would average it away with the other clusters. As
Figure~\subfigref{fig:granularity}{b} shows, IBCS thereby halves the recall gap of mean
pooling to the oracle top-$k$.

\subsection{Sampling with Inclusion-Probability Weighting (SIP)}
\label{sec:sip}

Top-$k$, threshold, and static-mask methods compute a fixed set of blocks and
simply drop all the rest from both the attention numerator and the softmax normalizer, a
loss that better scores can shrink but never fully remove. Attention mass, however, has a long
tail: as Figure~\subfigref{fig:granularity}{a} shows, even the oracle top-$k$ keeps only
about 69\% of it with 10.8\% of the keys. Recent correction methods~\citep{liu2025rectifiedspaattn,li2026pisa,zhou2026svgear}
approximate
the contribution of dropped blocks with an additional correction path. SIP instead
uses the block scores only to construct inclusion probabilities, computes every
sampled block exactly in a single sparse attention pass, and reweights these
contributions by inverse inclusion probability. Thus, beyond selection and sampling,
SIP requires no additional attention computation.

\paragraph{Score-proportional sampling.}
For each query block $Q_i$, SIP samples $s$ of the $M$ key/value blocks, with
$s$ set by the compute budget. Block $j$ is included with probability
\begin{equation}
  \pi_{ij}=\min(1,\lambda_iS_{ij}),\qquad \textstyle\sum_{j=1}^{M}\pi_{ij}=s,
  \label{eq:inclusion_prob}
\end{equation}
where $\lambda_i$ scales the scores to the budget. High-scoring blocks with
$\pi_{ij}=1$ are always kept, as in top-$k$, while every other block has a
nonzero inclusion probability required by the estimator below. SIP thus approaches
top-$k$ when the scores are concentrated and uniform sampling when they are
flat. It draws the sample $J_i$ by systematic
sampling~\citep{madow1949systematic}, which includes each block with
probability $\pi_{ij}$ and returns exactly $s$ distinct blocks, so the sample
fills whole tiles. All query tokens of $Q_i$ share $J_i$, so the sample
executes as block-sparse attention. Appendix~\ref{app:sampling} details both
steps.

\paragraph{Inverse-inclusion-probability weighting.}
Let $A_{ij}=\exp(Q_iK_j^\top/\sqrt d)$. Block $j$ contributes
$N_{ij}=A_{ij}V_j$ to the attention numerator and the row sums $D_{ij}$ of
$A_{ij}$ to the softmax normalizer. Dense attention sums both over all blocks,
$N_i=\sum_{j=1}^{M}N_{ij}$ and $D_i=\sum_{j=1}^{M}D_{ij}$, and computes
$O_i=N_i/D_i$ with row-wise division. Attending only to the sampled blocks would drop every
unsampled block from both sums, so the sample would again prune attention
rather than estimate it. SIP therefore weights each sampled block by
$1/\pi_{ij}$, so that it also stands in for the blocks left out:
\begin{equation}
  \widehat O_i=\frac{\widehat N_i}{\widehat D_i}
  =\frac{\sum_{j\in J_i}N_{ij}/\pi_{ij}}{\sum_{j\in J_i}D_{ij}/\pi_{ij}}
  =\operatorname{softmax}\!\Bigl(\frac{Q_iK_{J_i}^\top}{\sqrt d}-\log\pi_{iJ_i}\Bigr)V_{J_i},
  \label{eq:sip_estimate}
\end{equation}
where $K_{J_i}$ and $V_{J_i}$ stack the sampled blocks. A block with
$\pi_{ij}=1$ keeps weight one, while a block sampled with probability $1/4$
counts four times. Since $\Pr(j\in J_i)=\pi_{ij}$, $\widehat N_i$ and
$\widehat D_i$ are unbiased Horvitz--Thompson
estimates~\citep{horvitz1952sampling} of $N_i$ and $D_i$ for fixed
activations, and their variance shrinks as the scores become proportional to
the attention mass of the blocks.
The ratio $\widehat O_i$ itself is biased in general, but as the last form shows, it
is ordinary softmax attention over the sampled blocks with the logits of block
$j$ shifted by $-\log\pi_{ij}$, so it runs in a single standard attention pass.
Appendix~\ref{app:sampled_estimator} gives the derivation. As
Figure~\subfigref{fig:granularity}{c} shows, explicitly accounting for the tail brings the
attention error of SIP 41\% below that of the oracle top-$k$ at the same block
size.

\subsection{Fine-Grained Selection with Tiled Execution}
\label{sec:descriptor_execution}

Selection granularity and execution efficiency favor different block sizes.
Large key/value blocks align naturally with execution tiles and use tensor
cores efficiently, but make selection coarse. Small blocks select important
keys more precisely, but are too small to serve as efficient matrix-multiply
tiles on their own. AB-Attention co-designs these two levels: it selects
small key/value blocks and packs the sampled blocks into full execution tiles.
At a fixed sampling budget, this preserves the number of tensor-core tile
computations while allowing finer-grained selection.

For each query block, a fused attention kernel gathers its sampled key/value
blocks directly into full tiles, applies the inverse-inclusion-probability
logit shifts, and performs the usual online softmax and value accumulation in
one pass. The kernel gathers sampled blocks by index and packs them on the
fly, so noncontiguous blocks require neither token reordering nor a materialized
reordered sequence.

The selector follows the same tiled design. It scores multiple query and
key/value blocks in batched matrix operations and aggregates their partial
results, avoiding a separate operation for every small candidate block.
Sampling plans are rebuilt every $C$ denoising steps and reused between
rebuilds, while attention is recomputed from the current activations at every
step.

\begin{algorithm}[t]
\caption{AB-Attention for one attention head at denoising step $t$}
\label{alg:ab_attention}
\begin{algorithmic}[1]
\Require $Q,K,V\in\mathbb R^{n\times d}$; block sizes $B_q$ and $b$;
         clusters $R$; budget $s$; plan cycle $C$
\If{$t$ is a plan-rebuild step} \Comment{every $C$ steps}
  \State $\bar K_j\gets\textsc{MeanPool}(K_j)$ for $j=1,\dots,M$
  \For{each query block $Q_i$}
    \State $(\bar Q_i,w_i)\gets\textsc{Cluster}(Q_i,R)$ \Comment{IBCS}
    \State $S_{ij}\gets$ score of block $K_j$ (Eq.~\ref{eq:block_score}), $j=1,\dots,M$
    \State $\pi_{ij}\gets$ inclusion probability of $K_j$ (Eq.~\ref{eq:inclusion_prob}), $j=1,\dots,M$
           \Comment{SIP}
    \State $J_i\gets$ sample $s$ distinct blocks with $\Pr(j\in J_i)=\pi_{ij}$
           \Comment{systematic sampling}
  \EndFor
\EndIf
\For{each query block $Q_i$} \Comment{recomputed at every step}
  \State gather $K_{J_i},V_{J_i}$ and pack them into full tiles
  \State $\widehat O_i\gets
         \operatorname{softmax}\!\bigl(Q_iK_{J_i}^\top/\sqrt d-\log\pi_{iJ_i}\bigr)V_{J_i}$
\EndFor
\State \Return $\widehat O$
\end{algorithmic}
\end{algorithm}

\section{Experiments}

\subsection{Experimental Setup}
\label{sec:evaluation_protocol}

\paragraph{Models and datasets.}
We evaluate three text-to-video diffusion transformers:
\mbox{Wan2.1-T2V-1.3B} and \mbox{Wan2.1-T2V-14B}~\citep{wan2025wan}, and
\mbox{HunyuanVideo-T2V-13B}~\citep{kong2024hunyuanvideo}. Every generated video is sampled at
720p resolution and contains 81 frames on Wan2.1 or 129 frames on
HunyuanVideo. Text prompts are the 600 prompts of the Penguin
Benchmark~\citep{kong2024hunyuanvideo}.

\paragraph{Metrics.}
We compare each sparse
video with the corresponding full-attention video from the same prompt and seed and report PSNR,
SSIM~\citep{wang2004ssim}, and LPIPS~\citep{zhang2018lpips}. To evaluate the
videos themselves, independent of the reference, we further report five
VBench~\citep{huang2024vbench} dimensions: aesthetic quality (AQ), background
consistency (BC), imaging quality (IQ), motion smoothness (MS), and subject
consistency (SC). For efficiency, sparsity is the share of video-attention
computation skipped in the sparse denoising steps, and speedup is the ratio of
dense to sparse end-to-end latency.

\paragraph{Baselines.}
We compare against three recent state-of-the-art training-free sparse attention methods.
RadialAttention~\citep{li2025radial} applies a static energy-decay mask,
whereas SVG2~\citep{yang2025svg2} and DFSAttn~\citep{hu2026dfsattn} build
their masks dynamically from the current queries and keys. We use the official
released code and the default hyperparameters of each baseline, as detailed in
Appendix~\ref{app:baselines}. The released DFSAttn implementation executes its masks
with a FlashAttention-2-based block-sparse kernel~\citep{guo2024blocksparse}. For a
common backend, we execute the masks of all three baselines with FlashInfer's
FlashAttention-3 block-sparse backend~\citep{ye2025flashinfer}; this is also the
backend used by the released RadialAttention and SVG2 implementations.

\paragraph{Implementation details.}
Each video is generated on NVIDIA H20 GPUs in BF16 precision with PyTorch~2.6
and CUDA~12.4. Dense attention, including the dense steps of every method,
uses FlashAttention-3~\citep{shah2024flashattention3}. Following prior
work~\citep{li2025radial,hu2026dfsattn}, all methods keep attention dense in
the first transformer block and during the first 10 of the 50 denoising steps.
AB-Attention uses a key/value block size of 4, a query block size of 128, and
eight query clusters per query block, and rebuilds its sampling plan every
four denoising steps. Its sampling budget is kept fixed for each model, resulting in an
89.16\% sparsity on both Wan2.1 models and 92.35\% on HunyuanVideo.
End-to-end latency is measured following the protocol in Appendix~\ref{app:latency}.

\subsection{Main Results}
\label{sec:main_results}

\begin{table*}[t]
  \caption{Quality and efficiency comparison of AB-Attention and baselines on
  \mbox{Wan2.1-T2V-1.3B}, \mbox{Wan2.1-T2V-14B}, and \mbox{HunyuanVideo-T2V-13B}.
  Shaded rows report the full-attention reference.}
  \label{tab:main_results}
  \small
  \setlength{\tabcolsep}{4pt}
  \renewcommand{\arraystretch}{1.08}
  \begin{center}
  \resizebox{0.98\textwidth}{!}{%
  \begin{tabular}{c|ccc|ccccc|ccc}
    \toprule
    \multirow{2}{*}{\textbf{Method}} & \multicolumn{3}{c|}{\textbf{Perceptual Metrics}} & \multicolumn{5}{c|}{\textbf{VBench}} & \multicolumn{3}{c}{\textbf{Efficiency}} \\
    \cmidrule(lr){2-4} \cmidrule(lr){5-9} \cmidrule(lr){10-12}
    & PSNR $\uparrow$ & SSIM $\uparrow$ & LPIPS $\downarrow$ & AQ $\uparrow$ & BC $\uparrow$ & IQ $\uparrow$ & MS $\uparrow$ & SC $\uparrow$ & Sparsity $\uparrow$ & Latency (s) $\downarrow$ & Speedup $\uparrow$ \\
    \midrule
    \rowcolor{black!6}\textbf{Wan2.1 1.3B} & $\infty$ & 1.0000 & 0.0000 & 59.11 & 95.96 & 65.48 & 98.99 & 95.96 & 0\% & 962.39 & $1.00\times$ \\
    Radial & 27.856 & 0.8672 & 0.1704 & 59.65 & 95.97 & 66.56 & 99.02 & 95.63 & 73.63\% & 549.91 & $1.75\times$ \\
    DFSAttn & 27.734 & 0.8508 & 0.2193 & 57.83 & 95.78 & 67.30 & 98.93 & 95.45 & 81.03\% & 548.75 & $1.75\times$ \\
    SVG2 & 30.021 & 0.8858 & 0.1576 & 54.59 & 94.82 & 58.25 & 98.90 & 93.84 & 80.12\% & 536.47 & $1.79\times$ \\
    \rowcolor{reportaccent!8}\textbf{AB-Attention} & \textbf{34.436} & \textbf{0.9328} & \textbf{0.0734} & 58.02 & 95.75 & 63.92 & 98.96 & 95.84 & \textbf{89.16\%} & 479.11 & $2.01\times$ \\
    \cmidrule{1-12}
    \rowcolor{black!6}\textbf{Wan2.1 14B} & $\infty$ & 1.0000 & 0.0000 & 59.39 & 96.32 & 66.84 & 99.16 & 96.18 & 0\% & 4743.52 & $1.00\times$ \\
    Radial & 26.696 & 0.8595 & 0.1708 & 59.43 & 96.19 & 66.56 & 99.14 & 95.90 & 73.63\% & 2907.31 & $1.63\times$ \\
    DFSAttn & 27.708 & 0.8784 & 0.1464 & 58.50 & 96.12 & 67.04 & 99.15 & 95.82 & 81.03\% & 2866.46 & $1.65\times$ \\
    SVG2 & 29.236 & 0.8950 & 0.1274 & 58.08 & 95.90 & 65.27 & 99.09 & 95.66 & 67.89\% & 3169.10 & $1.50\times$ \\
    \rowcolor{reportaccent!8}\textbf{AB-Attention} & \textbf{30.788} & \textbf{0.9169} & \textbf{0.0964} & 58.92 & 96.23 & 66.09 & 99.14 & 96.11 & \textbf{89.16\%} & 2571.52 & $1.84\times$ \\
    \cmidrule{1-12}
    \rowcolor{black!6}\textbf{HunyuanVideo 13B} & $\infty$ & 1.0000 & 0.0000 & 57.45 & 96.12 & 66.13 & 99.16 & 95.39 & 0\% & 4399.03 & $1.00\times$ \\
    Radial & 27.686 & 0.8802 & 0.1655 & 57.21 & 96.04 & 65.89 & 99.16 & 95.07 & 73.03\% & 2412.27 & $1.82\times$ \\
    DFSAttn & 30.251 & 0.9141 & 0.1151 & 57.22 & 96.04 & 65.90 & 99.14 & 95.27 & 81.29\% & 2129.13 & $2.07\times$ \\
    SVG2 & 30.230 & 0.8967 & 0.1529 & 54.97 & 95.66 & 62.18 & 99.07 & 94.84 & 72.37\% & 2464.84 & $1.78\times$ \\
    \rowcolor{reportaccent!8}\textbf{AB-Attention} & \textbf{34.347} & \textbf{0.9480} & \textbf{0.0700} & 57.03 & 96.07 & 65.29 & 99.15 & 95.36 & \textbf{92.35\%} & 1974.61 & $2.23\times$ \\
    \bottomrule
  \end{tabular}}
  \end{center}
\end{table*}

\begin{figure}[p]
  \centering
  \includegraphics[width=\linewidth]{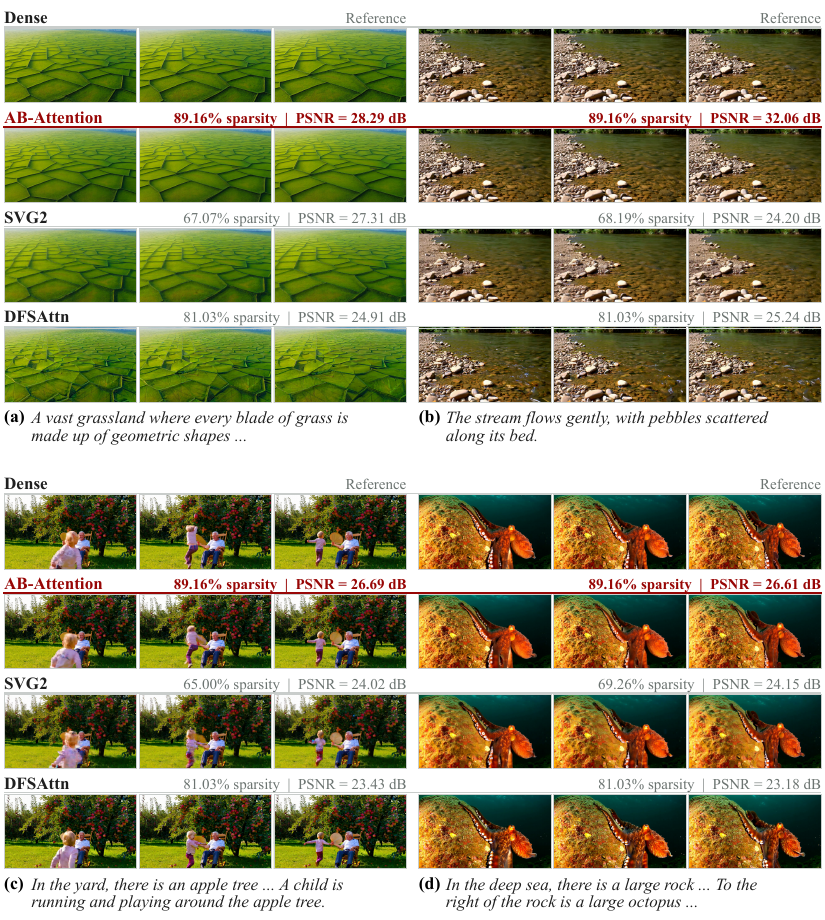}
  \caption{Qualitative comparison on \mbox{Wan2.1-T2V-14B}. Panels (a)--(d) are
  four Penguin Benchmark prompts, abbreviated with ellipses. Within a panel, rows are
  methods and columns are frames 12, 40, and 68 of the same video. Headers show sparsity and full-video PSNR relative to the dense reference.}
  \label{fig:qualitative}
\end{figure}

Table~\ref{tab:main_results} reports the quality and efficiency of every method on all three
models. AB-Attention attains the best PSNR, SSIM, and LPIPS of any method on
every model and consistently improves PSNR over the strongest baseline by 1.6--4.4~dB. It
does so while skipping 89--92\% of attention computation, more than any
baseline: every competing baseline computes 1.8--3.6 times as much attention.
Figures~\ref{fig:teaser} and~\ref{fig:qualitative} show the same effect
visually: AB-Attention stays close to the dense reference where the baselines
lose fine detail. On VBench, AB-Attention remains close to dense attention on all
five dimensions, showing that the videos it generates retain their overall
visual quality.

Efficiency follows the same ordering. AB-Attention reaches end-to-end speedups
of 1.84--2.23$\times$, ahead of every baseline on every model, and cuts the
generation time of a HunyuanVideo video from 4399~s to 1975~s. Its closest
competitor, DFSAttn on HunyuanVideo, reaches 2.07$\times$ only while computing
2.4 times as much attention, 18.71\% versus 7.65\% of dense attention.
AB-Attention is thus at once the sparsest, the
most faithful, and the fastest method on every model in
Table~\ref{tab:main_results}.

\subsection{Ablation Study}
\label{sec:component_diagnostic}

Table~\ref{tab:component_diagnostic} ablates the three core designs of AB-Attention
one at a time, keeping the other two fixed: the query clusters of IBCS, the
sampling and weighting of SIP, and the fine-grained key/value selection unit. All
variants use Wan2.1-T2V-14B with 100 prompts randomly sampled from the 600
Penguin Benchmark prompts and run at the same 89.16\% sparsity as the main results.

\begin{table}[t]
  \centering
  \begin{minipage}[b]{0.59\linewidth}
    \centering
    \caption{Ablation of the AB-Attention components on \mbox{Wan2.1-T2V-14B}.
    Each variant replaces one component at the same 89.16\% sparsity.}
    \label{tab:component_diagnostic}
    \vspace{4pt}
    \small
    \setlength{\tabcolsep}{3pt}
    \renewcommand{\arraystretch}{1.08}
    \newcommand{\dn}[1]{{\scriptsize\textcolor{red!65!black}{$-$#1}}}
    \newcommand{\nodn}{{\scriptsize\phantom{$-$0.00}}}
    \begin{tabular}[b]{llrlcc}
      \toprule
      & Variant & \multicolumn{2}{c}{PSNR $\uparrow$} & SSIM $\uparrow$ & LPIPS $\downarrow$ \\
      \midrule
      \rowcolor{reportaccent!8} & \textbf{AB-Attention} & \textbf{29.705} & \nodn & \textbf{0.9154} & \textbf{0.0956} \\
      \midrule
      IBCS & Query pooling & 28.793 & \dn{0.91} & 0.9024 & 0.1112 \\
      \cmidrule{1-6}
      \multirow{3}{*}{SIP} & w/o weights & 25.706 & \dn{4.00} & 0.8427 & 0.1911 \\
        & Uniform w/o weights & 22.452 & \dn{7.25} & 0.7437 & 0.4749 \\
        & Top-$k$ selection & 25.711 & \dn{3.99} & 0.8420 & 0.1996 \\
      \cmidrule{1-6}
      Tiling & Tile-level selection & 28.686 & \dn{1.02} & 0.9006 & 0.1113 \\
      \bottomrule
    \end{tabular}
  \end{minipage}\hfill
  \begin{minipage}[b]{0.39\linewidth}
    \centering
    \includegraphics[width=\linewidth]{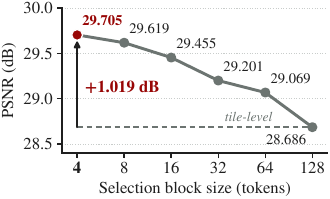}
    \setlength{\abovecaptionskip}{4pt}
    \captionof{figure}{PSNR versus selection block size on
    \mbox{Wan2.1-T2V-14B}.}
    \label{fig:block_size_psnr}
  \end{minipage}
\end{table}

\paragraph{Query clustering.}
IBCS summarizes each query block with its eight query clusters. To test whether
grouping queries by feature similarity matters, we instead average eight consecutive
16-query intervals, as in positional sub-block
pooling~\citep{hu2026dfsattn}, keeping the number of representatives and hence
the overall scoring cost unchanged. This \emph{query pooling} variant lowers PSNR by
0.91~dB, with SSIM and LPIPS degrading accordingly. A positional interval
mixes queries that attend to different keys, so its single mean can underrate a key
block that only some of them attend to, whereas a cluster gathers queries with
similar features, which tend to attend to similar keys.

\paragraph{Sampling and weighting.}
We compare SIP with three simpler alternatives at the same budget: score-proportional
sampling without the inverse-inclusion weights, uniform sampling without
weights, and deterministic top-$k$ selection. Removing the weights lowers PSNR
by 4.00~dB, to the level of top-$k$ selection: without them, the blocks
outside the sample are simply dropped, and sampling becomes another form of
pruning. Discarding the scores as well then costs a further 3.25~dB and more than
doubles LPIPS, because uniform sampling spends most of its fixed budget on the many
blocks that receive little attention, as Figure~\subfigref{fig:granularity}{a}
shows. The
scores thus decide where to spend the limited budget, and the weights account
for the blocks it leaves out; SIP needs both.

\paragraph{Selection granularity.}
\emph{Tile-level selection}, which scores and samples whole execution tiles as
conventional block-sparse attention does, lowers PSNR by 1.02~dB. As
Figure~\ref{fig:block_size_psnr} shows, PSNR rises steadily as the selection
block shrinks from a full tile toward the 4-token blocks of AB-Attention, and
Appendix~\ref{app:component_results} shows the same trend for SSIM and LPIPS.
Smaller key/value blocks let the sample keep the few important keys in a region without
spending any budget on their neighbors, and since the kernel of
Section~\ref{sec:descriptor_execution} packs them into full tiles, this gain costs no extra
tensor-core work.

Overall, replacing SIP costs by far the most, 3.99--7.25~dB of PSNR, followed
by tile-level selection at 1.02~dB and query pooling at 0.91~dB. Every variant
is worse than AB-Attention on all three metrics, so at the same budget each
component of AB-Attention contributes to its fidelity.

\section{Related Work}
\label{sec:related_work}

\paragraph{Sparse Attention.}
Sparse attention accelerates video diffusion transformers~\citep{peebles2022dit,ma2024latte,yang2024cogvideox} by restricting
query--key interactions to a subset that can be computed efficiently~\citep{child2019sparse,beltagy2020longformer,zaheer2020bigbird,roy2020routing,choromanski2020performer,xiong2021nystromformer}.
Static methods exploit video geometry with fixed sparsity patterns~\citep{chen2026sparsevdit}: Sliding
Tile Attention~\citep{zhang2025sta} uses tiled local 3D windows, Radial
Attention~\citep{li2025radial} constructs a static multiscale mask from
spatiotemporal energy decay, and Sparse VideoGen~\citep{xi2025sparsevideogen}
profiles attention heads online to assign each head a predefined spatial or
temporal pattern. Dynamic methods instead infer sparsity from the current
activations. XAttention~\citep{xu2025xattention} estimates block importance by summing
attention scores along strided antidiagonals, and SpargeAttention~\citep{zhang2025spargeattention}
combines coarse prediction from mean-compressed self-similar blocks with
softmax-aware filtering. To improve block
representations, SVG2~\citep{yang2025svg2} clusters tokens by semantic
similarity with $k$-means and permutes each cluster into contiguous blocks, and
DFSAttn~\citep{hu2026dfsattn} combines 3D Hilbert-curve reordering with
hierarchical sub-block scoring.
SPADE~\citep{liu2026spade} adapts block summaries and sparsity policies to
each input. AdaSpa~\citep{xia2025adaspa} and RAPID~\citep{lin2026rapid}
amortize sparsity search by reusing and adapting masks across denoising
steps.
Across these designs, selection granularity, grouping overhead, and
hardware-friendly execution remain coupled.

\paragraph{Correction-Based Sparse Attention.}
Pruning removes the skipped interactions from the attention numerator
and the softmax normalizer. Training-free correction methods approximate these
missing contributions: Rectified SpaAttn~\citep{liu2025rectifiedspaattn}
rectifies the amplified weights of retained blocks and recovers those of pruned
blocks from pooled attention,
PISA~\citep{li2026pisa} approximates the remaining blocks with a block-wise
Taylor expansion, and
SVG-EAR~\citep{zhou2026svgear} compensates pruned blocks with cluster
centroids and routes exact computation to the blocks with the largest
error. All three compute selected blocks
exactly and approximate the rest, whereas AB-Attention estimates both the
numerator and the normalizer from one weighted sample of all blocks.

\paragraph{Other Acceleration Methods.}
Diffusion inference can also be accelerated by reducing denoising steps~\citep{song2020ddim,lu2022dpmsolverpp,zhao2023unipc,song2023consistency,luo2023lcm},
reusing intermediate computation~\citep{ma2023deepcache,zhao2024pab,lv2024fastercache}, or reducing the sequence length processed per step~\citep{bolya2022tome,bolya2023tomesd}. Fast solvers such as DPM-Solver~\citep{lu2022dpmsolver} reduce the
number of model evaluations, while progressive
distillation~\citep{salimans2022progressive} repeatedly trains a student to
halve the number of sampling steps. TeaCache~\citep{liu2025teacache}
reuses cached model outputs across denoising steps while the accumulated
difference of timestep-modulated inputs stays below a threshold. DyDiT~\citep{zhao2025dydit} dynamically adjusts model width across denoising timesteps and skips computation for less important spatial tokens; DyDiT++~\citep{zhao2026dyditpp} extends this approach to video generation and other visual-generation settings. RAPID$^3$~\citep{zhao2026rapidthree} jointly learns policies for denoising-step skipping, feature-cache reuse, and sparse-attention selection. FlashAttention~\citep{dao2022flashattention} reduces
memory traffic by computing exact attention tile by tile in on-chip memory, and
SageAttention~\citep{zhang2024sageattention} accelerates attention by
quantizing queries and keys to INT8. These approaches are complementary to sparse
attention and can be combined with it~\citep{dao2023flashattention2,shah2024flashattention3}.

\section{Conclusion}
\label{sec:conclusion}

We presented AB-Attention, a training-free sparse attention method for video
diffusion transformers that directly estimates dense attention rather than truncating
it. IBCS scores key/value blocks with lightweight query clusters formed only within each query
block, and SIP samples blocks in proportion to these scores and reweights them
by their inverse inclusion probabilities, so that a single block-sparse pass
yields unbiased estimates of the attention numerator and normalizer.
AB-Attention further decouples selection from execution: it selects key/value
blocks much smaller than a tile and packs them into full tensor-core tiles in its kernel. On both Wan2.1 and HunyuanVideo, AB-Attention skips 89--92\% of attention, whereas the
baselines compute 1.8--3.6$\times$ as much, yet it stays closest to dense
attention and reaches the highest end-to-end speedups of 1.84--2.23$\times$.

\paragraph{Limitations and future work.}
\label{sec:limitations}
Despite these encouraging results, AB-Attention remains limited in both its
current implementation and application scope. Our kernel is built for the H20
GPUs used in our experiments and currently supports only forward inference,
while our evaluation covers three text-to-video diffusion transformers. Future
work will address these limitations through kernel support for further
hardware, evaluation on a broader range of models, and a trainable
implementation with backward support that enables fine-tuning under AB-Attention.

\clearpage
\phantomsection
\addcontentsline{toc}{section}{References}
\bibliography{custom}

\newpage
\appendix
\onecolumn
\section{Sampling and Estimation Details}
\label{app:details}

\subsection{Query Centroids and Pooled Keys}
\label{app:representatives}

The single $k$-means iteration of IBCS starts from $R$ centers chosen in each
query block by farthest-first traversal~\citep{gonzalez1985farthest}. For the
four-token key blocks, the first two and the last two keys are mean-pooled
into the two pooled keys of $\bar K_j$ in Equation~\ref{eq:block_score}.
Pooling halves the scoring cost, and since every pooled key stands for two
keys, key-side size weights cancel in the softmax.

\subsection{Systematic Sampling Without Replacement}
\label{app:sampling}

\paragraph{Inclusion probabilities.}
The scale $\lambda_i$ in Equation~\ref{eq:inclusion_prob} is found
iteratively: scale the remaining scores to the remaining budget, set
$\pi_{ij}=1$ for blocks whose scaled scores reach one, deduct them from the
budget, and repeat. Scores are floored at $10^{-30}$, so every $\pi_{ij}$
stays positive, as Appendix~\ref{app:sampled_estimator} requires. The budget
$s$ is a multiple of 32, so the sample fills whole 128-token tiles.

\paragraph{Systematic draw.}
Let $H_i=\{j:\pi_{ij}=1\}$ and $s'=s-|H_i|$. Order the remaining blocks as
$j_1,\ldots,j_L$ by index, let $F_0=0$ and
$F_t=\sum_{h=1}^{t}\pi_{i j_h}$, so $F_L=s'$, draw
$\xi\sim\operatorname{Uniform}[0,1)$, and include $j_t$ if
\begin{equation}
  F_{t-1}\le\xi+r<F_t
  \quad\text{for some }r\in\{0,\ldots,s'-1\}.
  \label{eq:systematic_draw}
\end{equation}
The intervals $[F_{t-1},F_t)$ tile $[0,s')$, which contains the $s'$
thresholds $\xi+r$ spaced one apart. An interval of length $\pi_{ij_t}<1$
contains at most one threshold, and does so with probability
$\pi_{ij_t}$~\citep{madow1949systematic}. With $H_i$, the draw
returns exactly $s$ distinct blocks with inclusion probabilities $\pi_{ij}$.
Independent per-block draws would make the sample size random, and
sampling with replacement could repeat blocks, leaving tiles partly empty or
wasting tensor-core work on duplicates.

\subsection{Weighted Attention Estimation}
\label{app:sampled_estimator}

With the block contributions $N_{ij}$ and $D_{ij}$ defined before
Equation~\ref{eq:sip_estimate}, let $I_{ij}$ indicate whether block $j$ is
sampled for query block $i$. The Horvitz--Thompson estimates
\begin{equation}
  \widehat N_i=\sum_{j=1}^{M}\frac{I_{ij}N_{ij}}{\pi_{ij}},
  \qquad
  \widehat D_i=\sum_{j=1}^{M}\frac{I_{ij}D_{ij}}{\pi_{ij}}
  \label{eq:block_ht}
\end{equation}
satisfy $\mathbb E[\widehat N_i]=N_i$ and $\mathbb E[\widehat D_i]=D_i$ for
fixed activations, since $\mathbb E[I_{ij}]=\pi_{ij}$; this needs no
independence between blocks, which systematic sampling lacks. The ratio
$\widehat O_i=\widehat N_i/\widehat D_i$ is biased in general. Since
$A_{ij}/\pi_{ij}=\exp(Q_iK_j^\top/\sqrt d-\log\pi_{ij})$, it is softmax
attention over sampled blocks with the logits of block $j$ shifted by
$-\log\pi_{ij}$, the last form of Equation~\ref{eq:sip_estimate}. Tiling the
sampled blocks changes neither their indices nor weights, so the
online-softmax kernel of Algorithm~\ref{alg:ab_attention} computes
$\widehat O_i$ exactly.

\section{Implementation Details}
\label{app:implementation}

\subsection{AB-Attention}
\label{app:runtime}

On Wan2.1, AB-Attention replaces the self-attention among video tokens, and
cross-attention to the text remains dense. HunyuanVideo attends jointly over
video and prompt tokens; AB-Attention samples only the video keys of video
queries, while attention to prompt keys and all attention from prompt queries
remain exact.

Both Wan2.1 models share one configuration. Queries are divided into blocks
of $B_q=128$ tokens and keys and values into blocks of $b=4$ tokens. IBCS
forms $R=8$ query clusters per query block with one $k$-means iteration from
farthest-first centers, and pools each key block into two keys, as
described in Appendix~\ref{app:representatives}. SIP samples $s=2048$ key/value blocks
(8192 keys) per query block on Wan2.1 and $s=2272$ (9088 keys) on
HunyuanVideo, 10.84\% and 7.65\% of the 18{,}900 and 29{,}700 blocks of a
video, which gives the sparsity of 89.16\% and 92.35\%. Attention stays
dense in the first transformer block and in denoising steps 1--10. From step
11, the sampling plan is rebuilt every $C=4$ steps, at steps 11, 15,
\ldots, 47, and reused for the steps in between. Sampling uses seed 42.

\subsection{Baselines}
\label{app:baselines}

RadialAttention~\citep{li2025radial}
uses a decay factor of 0.2 on Wan2.1 and 0.95 on HunyuanVideo, with the 720p
video tokens zero-padded to a multiple of its 128-token mask block.
SVG2~\citep{yang2025svg2} groups queries and keys into 100 and 500 clusters
and keeps key clusters up to a cumulative attention of 0.9, with a minimum
key ratio of 0.1.
DFSAttn~\citep{hu2026dfsattn} reorders tokens along a 3D Hilbert curve,
scores blocks with 16-token sub-blocks, and refreshes its mask every 12
steps; the retained block fraction starts at 0.3 and drops by 0.1 per refresh
down to 0.1.

\section{Evaluation Protocol}
\label{app:evaluation}

\paragraph{Quality metrics.}
PSNR, SSIM, and LPIPS with the VGG backbone are computed
on RGB frames in $[0,1]$, averaged over the frames of each video, and then
averaged over prompts. The five VBench dimensions are evaluated per video and
averaged over prompts.

\paragraph{Sparsity.}
Sparsity counts only the sparse denoising steps. It comes from the static
mask for RadialAttention, from the allocated top-$k$ budget for
DFSAttn, from the token pairs kept by the recorded cluster masks for SVG2, and from the recorded block selections for AB-Attention.

\begin{table}[htbp]
  \centering
  \caption{Fidelity to dense attention for each selection block size on
  \mbox{Wan2.1-T2V-14B} (100 prompts). Every row executes 8192 key tokens per
  query block in 128-token tiles (89.16\% sparsity). Block size 4 is the full
  AB-Attention, and block size 128 is tile-level selection.}
  \label{tab:kv_block_ablation}
  \small
  \begin{center}
  \begin{tabular}{cccc}
    \toprule
    Block size & PSNR $\uparrow$ & SSIM $\uparrow$ & LPIPS $\downarrow$ \\
    \midrule
    4   & \textbf{29.705} & \textbf{0.9154} & \textbf{0.0956} \\
    8   & 29.619 & 0.9150 & \textbf{0.0956} \\
    16  & 29.455 & 0.9108 & 0.0985 \\
    32  & 29.201 & 0.9092 & 0.1015 \\
    64  & 29.069 & 0.9072 & 0.1032 \\
    128 & 28.686 & 0.9006 & 0.1113 \\
    \bottomrule
  \end{tabular}
  \end{center}
\end{table}

\paragraph{Latency.}
\label{app:latency}
Latency is measured end to end for one video on NVIDIA H20 GPUs. The
timed region covers text encoding, all 50 denoising steps, and VAE decoding,
and excludes model loading, initial-noise loading, and video export. For
AB-Attention, plan construction and sampling fall inside the denoising steps
and are therefore timed.

\section{Additional Ablation Results}
\label{app:component_results}

Table~\ref{tab:kv_block_ablation} reports PSNR, SSIM, and LPIPS for the
block-size sweep of Figure~\ref{fig:block_size_psnr}; none of them degrades as
the selection block shrinks from 128 to 4 tokens.

\end{document}